\documentclass[10pt,journal,compsoc]{IEEEtran}
\ifCLASSOPTIONcompsoc
  \usepackage[nocompress]{cite}
\else
  \usepackage{cite}
\fi
\ifCLASSINFOpdf
\else
\fi

\usepackage[colorlinks,linkcolor=blue]{hyperref}
\usepackage{graphicx}
\usepackage{amsmath}
\usepackage{amssymb}
\usepackage{mathrsfs}
\usepackage{enumitem}
\usepackage[colorlinks,linkcolor=blue]{hyperref}
\usepackage{colortbl}
\usepackage{tabularx}
\usepackage{subfigure}
\usepackage{color}
\usepackage{bm}
\usepackage{bbm}
\usepackage{dsfont}
\usepackage{multirow}
\usepackage{booktabs}

\usepackage{hyperref}       
\usepackage{url}            
\usepackage{booktabs}       
\usepackage{amsfonts}       
\usepackage{nicefrac}       
\usepackage{microtype}      
\usepackage{graphicx}
\newlength\savewidth

\newcommand{\tbf}[1]{\textbf{#1}}
\newcommand{\udl}[1]{\underline{#1}}

\newcommand{\ph}[1]{{\textcolor{red}}}

\begin{document}

\title{Video Frame Interpolation with Many-to-many Splatting and Spatial Selective Refinement}

\author{
Ping Hu,
~Simon Niklaus,
~Lu Zhang,
~Stan Sclaroff,
~Kate Saenko
\IEEEcompsocitemizethanks{
\IEEEcompsocthanksitem P. Hu is with the School of Computer Science and Engineering, University of Electronic Science and Technology of China, Sichuan 611731 China, and the Department of Computer Science, Boston University, Boston, MA 02215 USA (e-mail: chinahuping@gmail.com).
\IEEEcompsocthanksitem S. Niklaus is with Adobe,  San Jose, CA 95110 USA (e-mail: simon.niklaus@outlook.com).
\IEEEcompsocthanksitem L. Zhang is with Dalian University of Technology,  Dalian 116024 China (e-mail: luzhang$\_$dut@mail.dlut.edu.cn).
\IEEEcompsocthanksitem S. Sclaroff is with the Department of Computer Science, Boston University, Boston, MA 02215 USA (e-mail: sclaroff@bu.edu).
\IEEEcompsocthanksitem K. Saenko is with the Department of Computer Science, Boston University, Boston, MA 02215 USA, and FAIR, Meta, Menlo Park, CA 94025 USA (e-mail: saenko@bu.edu). 
}
}


\IEEEtitleabstractindextext{%
\begin{abstract}
In this work, we first propose a fully differentiable Many-to-Many (M2M) splatting framework to interpolate frames efficiently. 
Given a frame pair, we estimate multiple bidirectional flows to directly forward warp the pixels to the desired time step before fusing any overlapping pixels. 
In doing so, each source pixel renders multiple target pixels and each target pixel can be synthesized from a larger area of visual context, establishing a many-to-many splatting scheme with robustness to undesirable artifacts. 
For each input frame pair, M2M has a minuscule computational overhead when interpolating an arbitrary number of in-between frames, hence achieving fast multi-frame interpolation. 
However, directly warping and fusing pixels in the intensity domain is sensitive to the quality of motion estimation and may suffer from less effective representation capacity. 
To  improve interpolation accuracy, we further extend an M2M++ framework by introducing a flexible Spatial Selective Refinement (SSR) component, which allows for trading computational efficiency for interpolation quality and vice versa.
Instead of refining the entire interpolated frame, SSR only processes difficult regions selected under the guidance of an estimated error map, thereby avoiding redundant computation.
Evaluation on multiple benchmark datasets shows that our method is able to improve the efficiency while maintaining competitive video interpolation quality, and it can be adjusted to use more or less compute as needed. 
\end{abstract}

}

\maketitle
\IEEEdisplaynontitleabstractindextext
\IEEEpeerreviewmaketitle

\IEEEraisesectionheading{\section{Introduction}}
\label{sec:intro}
\IEEEPARstart{V}IDEO frame interpolation (VFI) aims to increase frame rates of videos by synthesizing intermediate frames in between the original ones~\cite{siyao2021deep,baker2011database,chen2023human}.
As a classic problem in video processing, VFI contributes to many practical applications, including slow-motion animation, video editing, video compression, etc~\cite{jiang2018super,meyer2018deep,wu2018video,zhang2019fast,hu2020dipnet}. 
In recent years, a plethora of techniques for video frame interpolation have been proposed~\cite{meyer2018phasenet,meyer2015phase,zhang2020video,liu2020enhanced,Yu_2021_ICCV,tulyakov2021time,reda2019unsupervised,argaw2022ltvfi,tulyakov2022time}.  However, frame interpolation remains an unsolved problem due to challenges like occlusions, blur, and large motion.

The referenced research can roughly be categorized into motion-free and motion-based, depending on whether or not cues like optical flow are incorporated~\cite{kroeger2016fast,sun2018pwc,hu2018recurrent,hu2018motion,hu2019motion}.
Motion-free models typically rely on kernel prediction~\cite{cheng2020video,ding2021cdfi,niklaus2021revisiting,peleg2019net} or spatio-temporal decoding~\cite{choi2020channel,choi2021motion,kalluri2020flavr}, which are effective but limited to interpolating frames at fixed time steps and their runtime increases linearly in the number of desired output frames.
On the other end of the spectrum, motion-based approaches establish dense correspondences between frames and apply warping to render the intermediate pixels.  

\begin{figure}[t]
    \centering
    \includegraphics[width=1\linewidth]{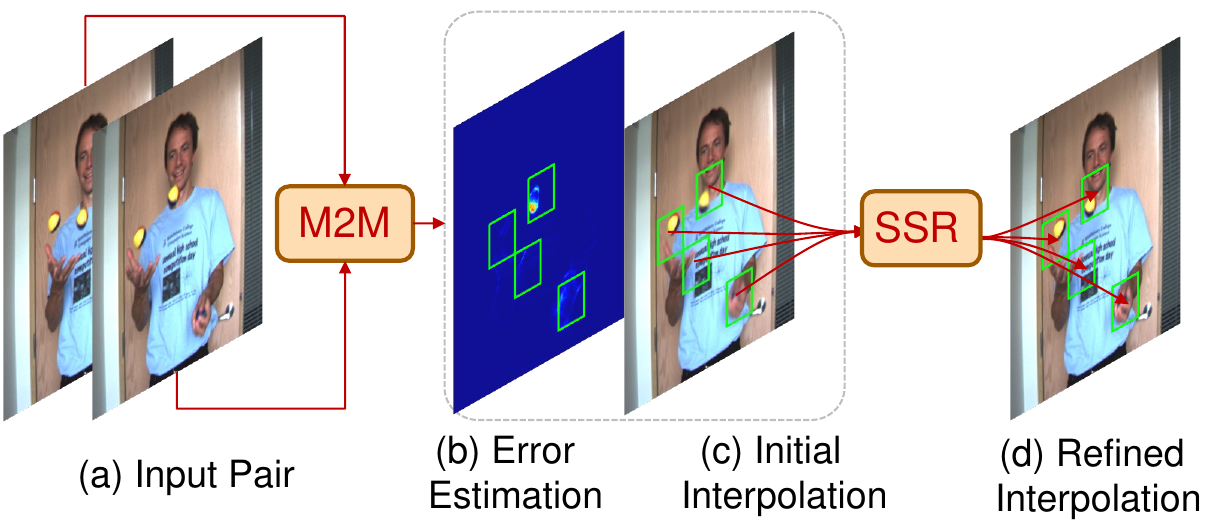}
    \vspace{-0.7cm}
    \caption{An overview of the proposed M2M++ framework for efficient video frame interpolation. Given an input pair (a), we first apply our many-to-many (M2M) splatting method to efficiently predict an error map (b) and an initial interpolation (c). A refined interpolation (d) is then generated by applying the spatial selective refinement (SSR) network to post-process challenging regions guided by the error map. By setting different thresholds for the erroneous region selection, M2M++ enables trading computational efficiency for interpolation quality and vice versa.}
    \label{fig:tesear}
    \vspace{-0.4cm}
\end{figure}

A common motion-based technique estimates bilateral flow for the desired time step and then synthesizes the intermediate frame via backward warping~\cite{huang2020rife,bao2019depth,park2021asymmetric,park2020bmbc,jiang2018super}. 
The estimation of bilateral motion is challenging though and incorrect flows can easily degrade the interpolation quality.
As a result,  for each time step, these methods typically apply a synthesis network to refine the bilateral flows.
Another motion-based solution is to forward warp pixels to the desired time step 
via optical flow~\cite{baker2011database}.
However, forward warping is subject to holes and ambiguities where multiple pixels map to the same location. 
Therefore, image refinement networks are commonly adopted to correct any remaining artifacts~\cite{niklaus2020softmax,xue2019video,niklaus2018context}. 
Both of these approaches require significant amounts of compute, and the refinement networks need to be executed for the entire frame at each of the desired interpolation instants.
This decreases their efficiency in multi-frame interpolation tasks since their run-time increases linearly in the number of desired output frames and their spatial resolutions.

\begin{figure}[t]
    \centering
    \includegraphics[width=0.95\linewidth]{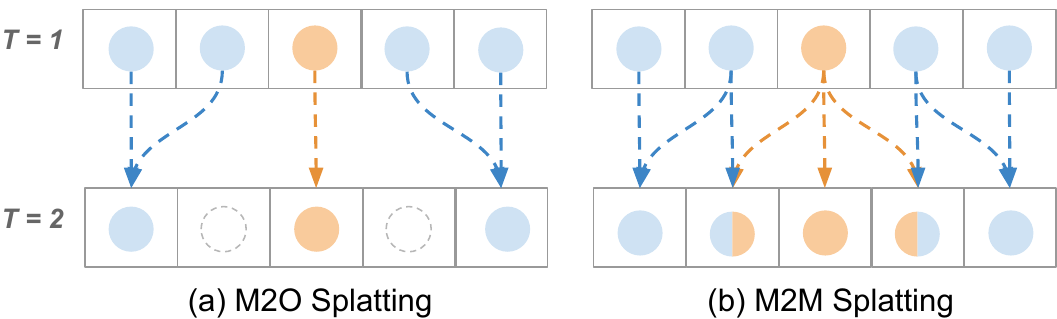}
    \vspace{-0.3cm}
    \caption{(a) Many-to-one splatting versus (b) many-to-many splatting for divergent flow in a scene containing blue and orange pixels. M2O splatting may results in holes, while M2M splatting allows for a more flexible image formation model.}
    \label{fig:warp}
    \vspace{-0.3cm}
\end{figure}
To tackle the above challenges and strive for efficiency, in~\cite{hu2022m2m}, we propose a Many-to-Many (M2M) splatting framework. M2M works by estimating multiple bidirectional flow fields and then efficiently forward warping the input images to the desired time step before fusing any overlapping pixels.
Since we directly operate on  pixel colors, the quality and resolution of the underlying optical flow play a critical role. 
For this reason, we first apply an off-the-shelf optical flow estimator~\cite{kroeger2016fast,sun2018pwc} to
extract the inter-frame motion between the two input frames at a coarse level.
Based on this low-resolution optical flow estimate, a Motion Refinement Network (MRN) predicts multiple flow vectors for each pixel at the full-resolution which we then use for our image synthesis through many-splatting.

Conventional motion-based frame interpolation methods only estimate one inter-frame motion vector for each pixel~\cite{niklaus2020softmax,niklaus2018context,xue2019video,huang2020rife,bao2019depth,park2021asymmetric,park2020bmbc}.
However as shown in Fig.~\ref{fig:warp} (a), forward warping with such a motion field manifests as many-to-one splatting, leaving unnecessary holes in the warped result.
To overcome this limitation, we model a many-to-many relationship among pixels by predicting multiple motion vectors for each of the input pixels, 
and then forward warp the pixels to multiple locations at the desired time step. 
As shown in Fig.~\ref{fig:warp} (b), many-to-many splatting allows for more complex interactions among pixels, i.e. each source pixel is allowed to render multiple target  pixels and each target pixel can be synthesized with a larger area of visual context.
Unsurprisingly, many-to-many splatting leads to many more overlapping pixels.  
To merge these, we further introduce a learning-based fusion strategy which adaptively combines pixels that map to the same location.
Since the optical flow estimation step  in our pipeline predicts time-invariant correspondence estimates, it only needs to be performed once for a given input frame pair. 
Once the many-to-many inter-frame motion has been established, generating new in-between frames only requires warping and fusing the input images. 

However, M2M~\cite{hu2022m2m} warps and fuses pixels directly in the RGB color space, which may suffer from less effective representation capacity, and the interpolation result is directly affected by the quality of motion estimation, which is likely inaccurate in areas with large or complex motion. 
To improve the interpolation quality in such challenging regions, in this paper we extend our previous framework~\cite{hu2022m2m} as M2M++ by further introducing an interpolation quality prediction mechanism and a Spatial Selective Refinement (SSR) framework to selectively improve the interpolated results with fine-grained refinement as illustrated in Fig.~\ref{fig:tesear}, thereby avoiding redundant computation in the entire spatial domain and
providing high flexibility for different speed-accuracy trade-offs.
In contrast to previous approaches~\cite{niklaus2020softmax,niklaus2018context} that leverage refinement networks over the full spatial domain of an interpolated frame, the proposed SSR only processes regions which are likely subject to artifacts that would benefit from additional post-processing. 
The region selection is guided by a learning-based error estimation map, where each interpolated pixel is associated with a score characterizing its reliability.
By ranking the error estimation scores and varying the threshold by which regions are selected, M2M++ is able to trade computational complexity for interpolation quality and vice versa.
We show that our proposed M2M++ is able to improve efficiency while maintaining competitive video interpolation quality at a state-of-the-art level.

In this paper, we first present our M2M splatting approach in Sec.~\ref{sec:m2m}. Then, we extend this framework as M2M++ by adding a Spatial Selective Refinement module in Sec.~\ref{sec:ssr}. The main contributions of this paper can be summarized as:

\begin{itemize}[itemsep=5pt,topsep=0pt,parsep=0pt]
    \item A Motion-Refinement Network (MRN) that estimates a many-to-many relationship between the two input images.
    \item A Many-to-Many (M2M) splatting synthesis model for very efficient arbitrary frame interpolation.
    \item A Spatial Selective Refinement (SSR) that post-processes challenging regions that are likely subject to artifacts in the initial interpolation. 
    \item A M2M++ framework that makes it possible to trade computational efficiency for interpolation quality and vice versa. Our experiments demonstrate that the proposed framework is able to significantly improve efficiency while maintaining competitive video interpolation quality.
\end{itemize}

\section{Related Work}
\label{sec:related}

Motion plays a key role for existing VFI methods, as it explicitly models pixel-level correspondence and trajectories across frames~\cite{zhang2020unsupervised,zhang2019capsal,yang2021learning,hu2020temporally,hu2023m2m}.
Motion-based video frame interpolation approaches typically estimate optical flows~\cite{kroeger2016fast,sun2018pwc} from given frames, and then propagate pixels/features  to the desired target time step~\cite{xue2019video,yuan2019zoom,zhang2020flexible,niklaus2022splatting}. 
Forward warping is an efficient solution to achieve this goal~\cite{baker2011database}. 
With bidirectional optical flow between given frames, Niklaus \textit{et al.}~\cite{niklaus2018context} directly forward warp the images as well as contextual features to the interpolation instant before utilizing a synthesis network to render the output frame. To make this z-buffered splatting fully differentiable, they further introduce softmax splatting~\cite{niklaus2020softmax} which allows them to train the feature extraction end-to-end.
Splatting has its downsides though, since it is not only necessary to address ambiguities of multiple pixels mapping to the same location but it is also necessary to handle the holes that are present in the sparse result.

To avoid having to handle these challenges, some methods are based on backward warping instead~\cite{bao2019memc,sim2021xvfi}. 
The necessary bilateral flow can, for example, be approximated from off-the-shelf flow estimates through a neural network~\cite{jiang2018super} or depth-based splatting~\cite{bao2019depth}. 
Park \textit{et al.}~\cite{park2020bmbc,park2021asymmetric} extend these ideas and introduce a network to further improve the motion representations while Huang~\textit{et al.}~\cite{huang2020rife} learn to directly estimate bilateral flows. Reda~\textit{et al.}~\cite{reda2022film} learn “scale-agnostic” motion estimator for generalization to both small and large motion.  
However, estimating bilateral flow is still challenging and the backward warped pixels may still suffer from artifacts. 
As a result, these methods also rely on image synthesis networks to improve the interpolation quality~\cite{huang2020rife,park2020bmbc,park2021asymmetric,niklaus2020softmax,niklaus2018context,lu2022vfitrans,kong2022ifrnet,danier2022mfnet}. Though being effective, the bilateral flow estimation and the image synthesis networks need to be fully executed for each desired output, leading to a linearly increasing runtime when interpolating more than one in-between frame.

In contrast to these methods, our approach relies on many-to-many splatting to alleviate some of the issues of splatting-based interpolation. Moreover, our M2M++ enables post-processing only specific regions that are likely prone to artifacts in the initial interpolation result, which makes it possible to trade computational efficiency for interpolation quality and vice versa.

Another dominant research direction for VFI aims to avoid explicit motion estimation altogether. One popular approach is to resample input pixels with spatially adaptive filters~\cite{liu2017video,peleg2019net}. Niklaus~\textit{et al.}~\cite{niklaus2017adaconv} estimate spatially-varying kernels which in subsequent work are decomposed into separable kernels~\cite{niklaus2017sepconv,niklaus2021revisiting}, which also formulate a many-to-many correlations between pixels. However, as local patches suffer from a limited spatial range, deformable convolutions are introduced to handle large motion~\cite{lee2020adacof,cheng2020video}.
To improve model efficiency, Ding~\textit{et al.}~\cite{ding2021cdfi} introduce model compression~\cite{lee2020adacof}.
Spatio-temporal decoding methods are also proposed to directly convert spatio-temporal features into target frames via channel attention~\cite{choi2020channel,choi2021motion} or 3D convolutions~\cite{kalluri2020flavr}.
However, most of these methods  generate outputs at a fixed time, typically halfway between the input images, which limits arbitrary-time interpolation and linearly increases the runtime for multi-frame interpolation.     

\begin{figure*}[t]
    \centering
    \includegraphics[width=\linewidth]{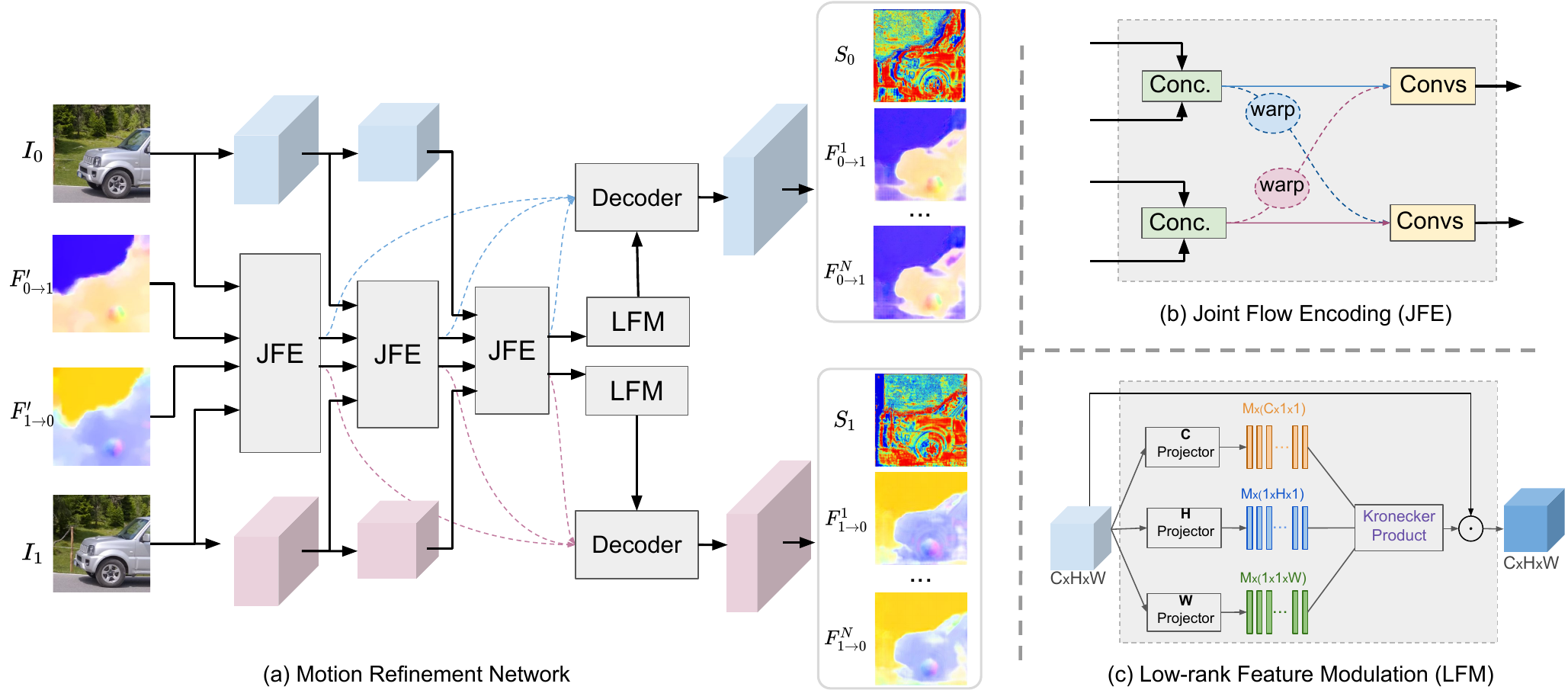}
    \vspace{-0.7cm}
    \caption{ Overview of the  (a) Motion Refinement Network and its core modules: (b) Joint Flow Encoding  and (c) Low-rank Feature Modulation. Given an image pair $\{I_0,I_1\}$ and the initial bidrectional inter-frame flow $\{F'_{0\rightarrow1}, F'_{1\rightarrow0}\}$, the goal is to generate multiple refined bidirectional flows $\{F^i_{0\rightarrow1}, F^i_{1\rightarrow0}\}_{i=1}^N$ and the reliability maps $\{S_{0},S_{1}\}$. The ``warp'' in the JFE denotes backward warping.}
    \label{fig:overview}
    \vspace{-0.4cm}
\end{figure*}
\section{Many-to-many Splatting}
\label{sec:m2m}
In this section, we describe our Many-to-Many (M2M) splatting framework for video frame interpolation. 
Given an input frame pair,  we first estimate the bidirectional motion with an off-the-shelf optical flow estimator~\cite{sun2018pwc,kroeger2016fast}. 
A Motion Refinement Network (Fig.~\ref{fig:overview} (a)) then takes the off-the-shelf motion predictions as input and estimates multiple motion vectors as well as a reliability score for each pixel.
Lastly, all input pixels are forward warped to the desired target time step several times via each of the multiple motion vectors, and finally merged to generate the output via a pixel fusion that leverages the estimated reliability score.
With full end-to-end supervision, our M2M framework achieves not only efficiency but also effectiveness.
In the following, we first present the Motion Refinement Network in Sec.~\ref{subsec:multi-flow}, then the multi-splatting and fusion of pixels in Sec.~\ref{subsec:m2msplatting}.

\subsection{Motion Refinement Network}
\label{subsec:multi-flow}
Optical flow is a common technique to model inter-frame motion in videos.
Yet, directly applying an off-the-shelf optical flow estimator and forward warping pixels based on this estimate may be challenging. 
Optical flow only models a single motion vector for each pixel, thus limiting the area that a pixel can splat to and thus potentially causing holes. 
Moreover, most optical flow estimators are supervised with training data at a relatively low resolution and forcing them to process high-resolution frames may yield poor results.
In contrast, we present a Motion Refinement Network (MRN) to upsample and refine an off-the-shelf optical flow estimate while predicting multiple motion vectors per pixel.
As shown in Fig.~\ref{fig:overview} (a), the MRN pipeline is composed of three parts: Motion Feature Encoding, Low-rank Feature Modulation, and Output Decoding.

\subsubsection{Motion Feature Encoding} 
Motion Feature Encoding aims to encode multi-stage motion features from the input frames $\{I_0, I_1\}$ and is guided by the optical flow $\{F'_{0\rightarrow1}, F'_{1\rightarrow0}\}$ from an off-the-shelf estimator~\cite{sun2018pwc,kroeger2016fast} at a coarse resolution.
As outlined in Fig.~\ref{fig:overview} (a), the encoding process is designed in a hierarchical manner.
At first, we extract two $L$-level image feature pyramids from $I_0$ and $I_1$, with the zeroth-level being the images themselves. 
To generate the feature representations at each pyramid level, we utilize two convolutional layers with PReLU activations to downsample the features from the previous level by a factor of two.
In our implementation, we use $L=4$, and the numbers of feature channels from shallow to deep are $16$, $32$, $64$, and $128$ respectively.

Then, from the zeroth to the last level, we apply Joint Flow Encoding (JFE) modules as illustrated in Fig.~\ref{fig:overview} (b) to progressively generate motion feature pyramids for the bidirectional flow fields $F'_{0\rightarrow1}$ and  $F'_{1\rightarrow0}$. 
In the $l$-th level's JFE module, the motion and image features from the previous level are warped towards each other. Specifically, the features from the pyramid corresponding to $I_0$ are warped towards $I_1$ and vice versa using the off-the-shelf optical flow estimates.
Then, the original features and the warped features are combined and downsampled using a two-layer CNN to encode the $l$-th level's motion features. 

\subsubsection{Low-rank Feature Modulation} 
Low-rank Feature Modulation is designed to further enhance the motion feature representations  with a low-rank constraint. 
The idea behind this module is that flow fields of natural dynamic scenes are highly structured due to the underlying physical constraints, which can be exploited by low-rank models to enhance the motion estimation quality~\cite{dong2014nonlocal,tang2020lsm,Lara_2016_CVPR,roberts2009learning,fleet2000design}.
To avoid formulating explicit optimization objectives like in previous methods, which may be inefficient in high-resolution applications, we draw inspirations from Canonical Polyadic (CP) decomposition~\cite{kolda2009tensor} and construct an efficient low-rank modulation component to enhance each flow's feature maps with low-rank characteristics.

As shown in Fig.~\ref{fig:overview} (c), given an input feature map of size $C\times H\times W$, three groups of projectors are adopted to respectively shrink the feature maps into the \textit{channel}, \textit{height}, and \textit{width} dimensions. 
Each projector is composed of a pooling layer, $1\times 1$ conv layers, and a sigmoid function. 
We apply $M$ projectors for each of the three dimensions which results in three groups of 1-D features, whose sizes can be represented as $M\times (C\times 1\times 1)$ for the \textit{channel} dimension,  $M\times (1\times H\times 1)$  for the \textit{height} dimension, and $M\times (1\times 1\times W)$ for the \textit{width} dimension.
Then, for each of the $M$ vectors from the three dimensions, we apply the \textit{Kronecker Product} to get a rank-1 tensor, whose shape is $C\times H\times W$.
The $M$ rank-1 tensors are later averaged point-wise. 
To ensure low-rank characteristic, $M$ is set to be smaller than $C$, $H$, and $W$ (we adopt $M=16$ in this work). 
We combine the input features and the low-rank tensor via point-wise multiplication, where the latter serves as weights to modulate the former with low-rank characteristics.

Deep learning-based low-rank constraints have also been utilized for model compression~\cite{phan2020stable}, segmentation~\cite{chen2020tensor} and  image reconstruction~\cite{zhang2021learning}. In this work we explore applying it to motion modeling and demonstrate its effectiveness on the task of video frame interpolation.

\subsubsection{Output Decoding}
\begin{figure}[t]
    \centering
    \includegraphics[width=\linewidth]{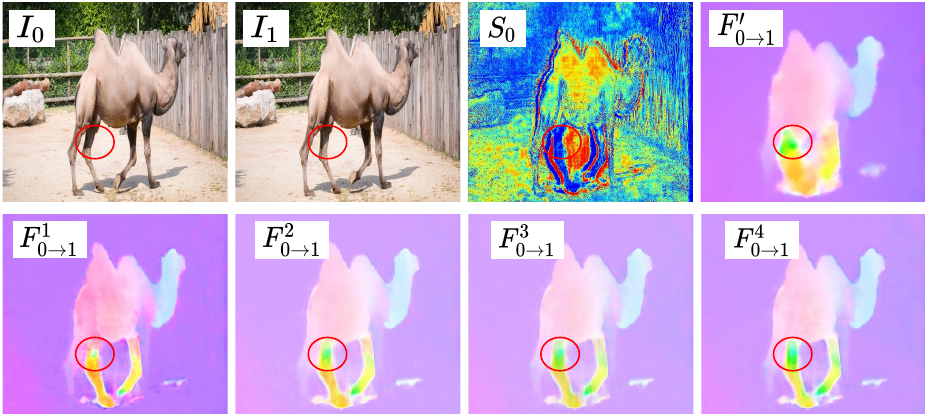}
    \vspace{-0.5cm}
    \caption{Examples of the MRN's output ($N=4$). $S_0$ shows low reliability (blue color) in areas with occlusion or smooth texture. $\{F_{0\rightarrow 1}^N\}_{n=1}^4$ refine the initial flow $F'_{0\rightarrow 1}$ with better details, and decompose complex motion with shade changes (indicated by the red circle) into multiple motion fields.
    }
    \label{fig:multi_flow}
    \vspace{-0.5cm}
\end{figure}

Output Decoding generates $N$ motion vectors as well as the reliability scores for each input pixel based on the motion feature pyramids and the feature maps subject to the low-rank prior. 
We adopt deconv layers to enlarge the spatial size of the feature maps.
That is, the decoder operates in $L$ stages from coarse to fine while leveraging the features encoded by the JFE modules.
At the last decoding stage, the full-resolution feature maps for the flow in each direction are converted into multiple fields $\{F^i_{0\rightarrow 1}, F^i_{1\rightarrow 0}\}_{i=1}^N$ as well as the corresponding reliability maps $\{S_{0}, S_{1}\}$, which are later utilized to fuse pixels that map to the same location when generating the new in-between frames. An example is visualized in Fig.~\ref{fig:multi_flow}.

\begin{figure*}[t]
    \centering
    \includegraphics[width=1\linewidth]{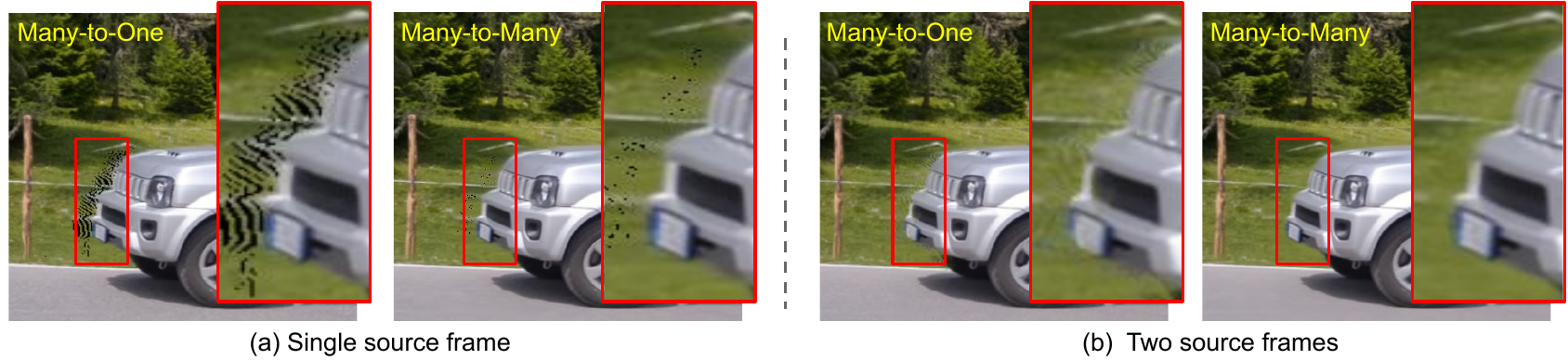}
    \vspace{-0.7cm}
    \caption{Visualization of forward warping via many-to-one (M2O)  splatting and many-to-many (M2M) splatting. (a) With one source frame, M2M splatting suffers less from banding artifacts and provides improved robustness to ambiguities near the boundaries of discontinuous motion. (b) Banding artifacts can be alleviated with multiple source frames, yet M2O splatting still suffers from stray effects at boundaries due to its image formation model that is less flexible than M2M splatting. Best viewed when zoomed in. }
    \vspace{-0.3cm}
    \label{fig:visual_warp}
\end{figure*}

\subsection{Pixel Warping and Fusion}
\label{subsec:m2msplatting}
The previously estimated multi-motion fields are first used to forward warp pixels to a given target time step. 
Later, we present a fusion strategy to combine the colors of overlapping pixels in the output. 
Since both the warping and fusion steps operate in color space without any subsequent post-processing steps, an intermediate frame can be interpolated with minuscule computational overhead.

\subsubsection{Pixel Warping.} So far, we have generated $N$ full-resolution bidirectional  motion fields $\{F^n_{0\rightarrow1},F^n_{1\rightarrow0}\}_{n=1}^N$ and pixel-wise reliability scores $\{S_0,S_1\}$ for the input video frame pair $\{I_0,I_1\}$.
The next step is to synthesize an intermediate frame $I_t$ at the desired time step $t\in(0,1)$.
Under the assumption of linear motion, we first scale each pixel's motion vectors by the desired interpolation time $t$ as:
\begin{equation}
\label{eq:flow2t}
\begin{aligned}
        F^n_{0\rightarrow t}(i_0) &= t \cdot F^n_{0\rightarrow 1}(i_0) \\
        F^n_{1\rightarrow t}(i_1) &= (1-t) \cdot F^n_{1\rightarrow 0}(i_1)
\end{aligned}
\end{equation}

\noindent where $i_0$ and $i_1$ denote the $i$-th source pixel in $I_0$ and $I_1$ respectively. Then, a source pixel $i_s$ is forward warped by its $n$-th motion vector to $i_{s\rightarrow t}^n = \phi_F(i_s,F^n_{s\rightarrow t})$ at the desired intermediate time $t$, with $s\in\{0,1\}$ representing the source frame, $\phi_F$ is the forward warping operation, and $F^n_{s\rightarrow t}$ is the $n$-th sub-motion vector of $i_s$ as defined in Eq.~\ref{eq:flow2t}.
    
We first consider utilizing a single motion vector for warping, which means each pixel is only warped to one location in the target frame. In dynamic scenes, the motion vectors may overlap with each other thus resulting in a many-to-one (M2O) propagation where the pixel set after fusion is smaller than the actual set of pixels in the frame. This results in holes as shown in Fig.~\ref{fig:visual_warp} (a). 
Though exploiting multiple source frames lessens this issue, M2O warping still restricts each source pixel to only render a small 4-pixel vicinity in the output frame.
This limits the effectiveness in representing and thus interpolating regions with complex interactions among the pixels, as shown in Fig.~\ref{fig:visual_warp} (b).

Fortunately, such limitations can be alleviated through many-to-many (M2M) pixel splatting by using multiple motion vectors to model the motion of each source pixel. 
We forward warp each pixel in the source $s$ with $N$ ($N>1$) sub-motion vectors to $t$, and get the set of warped pixels $\hat{I}_{s\rightarrow t}$,
\begin{align}
\label{eq:warpcolor}
\hat{I}_{s\rightarrow t} = \bigcup\limits_{n=1}^N\hat{I}^n_{s\rightarrow t}
\end{align}
where $\hat{I}^n_{s\rightarrow t}$ represents the frame warped from the source $s$ via the $n$-th motion fields.

Many-to-many splatting relaxes the restriction that each source pixel can only contribute to a single location. 
Therefore it allows the underlying motion estimator to learn to reason about occlusions, and model complex color interactions across a larger area of pixels. 
 
\subsubsection{Pixel Fusion.}
By applying M2M warping to all the input pixels in $\{I_0, I_1\}$, we get the complete warped pixel set where multiple target pixels may correspond to the same pixel locations:  $\hat{I}_t =\hat{I}_{0\rightarrow t}\bigcup\hat{I}_{1\rightarrow t}$.
To fuse warped pixels that overlap with each other, we measure each of the pixels' importance from three aspects: the temporal relevance, brightness consistency, and the reliability score.

\textit{1) Temporal Relevance} r{$_i$} characterizes changes not based on motion (e.g. lighting changes)  between a source frame and the target. 
For simplicity, we adopt linear interpolation by setting $r_i=1-t$ if $i$ comes from $I_0$ and $r_i=t$ otherwise, with $t$ being the desired interpolation time. 

\textit{2) Brightness Consistency} b{$_i$} indicates occlusions by comparing a frame to its target through backward warping:
\begin{align}
b_i=
\begin{cases}
-1\cdot||I_0(i)-I_1(i+F_{0\rightarrow 1}(i))||_1, &\text{\hspace{-0.2cm}if}~~i\in I_0,\\
-1\cdot||I_1(i)-I_0(i+F_{1\rightarrow 0}(i))||_1, &\text{\hspace{-0.2cm}if}~~i\in I_1,
\end{cases}
\label{eqn:occ}
\end{align}
where $F_{0\rightarrow 1}$ and $F_{1\rightarrow 0}$ are the averaged motion fields.

The effectiveness of Eq.~\ref{eqn:occ} is not decided only by the motion but also by the pixels' colors, which can be affected by various factors like noise, ambiguous appearance, and changes in shading~\cite{niklaus2020softmax,baker2011database}. To enhance the robustness, we thus further adopt a learned per-pixel reliability score.

\textit{3) Reliability Score} s$_i$ is jointly estimated together with the motion vectors through the  Motion Refinement Network as introduced in Sec.~\ref{subsec:multi-flow} and learned from data. 

With these three measurements, we fuse the overlapped pixels at a location $j$ in the form of weighted summation,
\begin{equation}
    I_t(j) = \frac{\sum_{i\in\hat{I}_t}\mathds{1}_{i=j}\cdot \mathbf{e}^{(b_{i}\cdot s_{i}\cdot\alpha)}\cdot r_{i}\cdot c_i}{\sum_{i\in\hat{I}_t}\mathds{1}_{i=j}\cdot  \mathbf{e}^{(b_{i}\cdot s_{i}\cdot\alpha)}\cdot r_{i}}
\label{eq:m2m}
\end{equation}
\noindent where $c_i$ represents the $i$-th warped pixel's original color, $\alpha$ is a learnable parameter adjusting the scale of weights, $\hat{I}_t$ is the set of all the warped pixels at time $t$, and $\mathds{1}_{i=j}$ indicates if the warped pixel  $i$ is mapping to the pixel location $j$. 

We note that our final fusion function is similar to SoftSplat~\cite{niklaus2020softmax} in the form of softmax weighting, however our method differs  in two aspects. First, we provide a solution to directly operate in the pixel color domain, while SoftSplat splats features and utilizes an image synthesis network instead. Second, we introduce the learning-based reliability score to fuse overlapping pixels in a data-driven manner while SoftSplat uses feature consistency.


\section{Spatial Selective Refinement}
\label{sec:ssr}
As M2M warps and fuses pixels directly in color space, noisy or blurry inputs as well as inaccurate inter-frame motion estimates significantly affect the interpolation quality. 
In this section, we address this limitation by extending the M2M framework with a Spatial Selective Refinement (SSR) module. Specifically, SSR refines the initial interpolation result to ameliorate possible artifacts. But instead of doing this naively on the initial interpolation result in its entirety, we employ the SSR only in regions that are likely subject to artifacts based on an error estimation step. This not only improves the computational efficiency overall by not refining areas that are already good, but it also makes it possible to trade computational efficiency for interpolation quality and vice versa by being more or less selective as to where to run the SSR. 
In the following, we will first introduce the estimation of per-pixel interpolation errors in Sec.~\ref{subsec:inter_error}, then the candidate patch selection in Sec.~\ref{subsec:cad_sel}, and finally present the patch-based refinement network in Sec.~\ref{subsec:refine_net}.

\subsection{Interpolation Error Prediction}
\label{subsec:inter_error}
As introduced in the previous section, when performing many-to-many splatting, each pixel in the target frame is interpolated by fusing input pixels forward-warped via bidirectional motion vectors. Therefore, the accuracy of an interpolated pixel is decided by the reliability of motion vectors associated with it.
Based on this, we modify the output of the previous Motion Refinement Network to additionally predict an error score which measures the reliability of each estimated motion vector. This error score can then be many-to-many splatted just like the input frames in order to get an error prediction map with respect to the interpolation result.

Formally, when estimating $N$  full-resolution bidirectional motion fields $\{F_{0\rightarrow 1}^n,F_{1\rightarrow 0}^n\}_{n=1}^N$, the Motion Refinement Network as introduced in the Sec.~\ref{subsec:multi-flow} also yields  $N$ pairs of full-resolution error score maps $\{E_{0\rightarrow 1}^n,E_{1\rightarrow 0}^n\}_{n=1}^N$ where each motion vector is associated with an error score. Each error score is normalized by a sigmoid function to ensure a value between 0 and 1. 
\begin{figure}[b]
    \centering
    \includegraphics[width=1\linewidth]{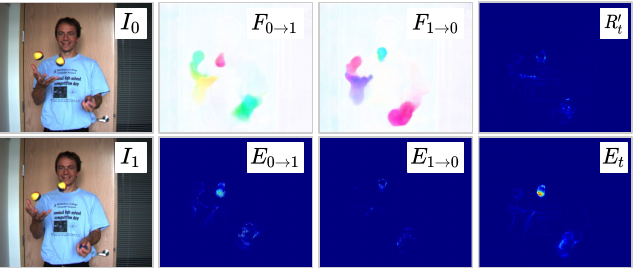}
    \vspace{-0.5cm}
    \caption{A visual example of error prediction step. $F_{0\rightarrow1}$ and $F_{1\rightarrow0}$ are the estimated motion fileds, $E_{0\rightarrow1}$ and $E_{1\rightarrow0}$ are the corresponding error score maps. As shown in the last column, the predicted error map ($E_t$) is aligned with the normalized residual map ($R'_t=\frac{R_t}{R^{Max}_t}$) between the interpolated frame and the groundtruth. Best viewed when zoomed in.}
    \label{fig:errorpred}
    \vspace{-0.5cm}
\end{figure}

Then, we forward-warp each source error score to the desired time step $t$ via its corresponding motion vector synthesized as in Eq.~\ref{eq:flow2t}. Note that multiple warped error scores may overlap at  the target pixel locations, like their corresponding pixel colors warped in Eq.~\ref{eq:warpcolor}.
Therefore, the error scores of a pixel location $j$ in the target frame can be fused just like the splatted input frames in Eq.~\ref{eq:m2m} as,
\begin{align}
\label{eq:errm2m}
    E_t(j) = \frac{\sum_{i\in\hat{E}_t}\mathbbm{1}_{i=j}\cdot \mathbf{e}^{(b_{i}\cdot s_{i}\cdot\alpha)}\cdot r_{i}\cdot e_i}{\sum_{i\in\hat{E}_t}\mathbbm{1}_{i=j}\cdot  \mathbf{e}^{(b_{i}\cdot s_{i}\cdot\alpha)}\cdot r_{i}}
\end{align}
where $e_i$ represents the $i$-th warped error score, $\hat{E}_t$ is the set of all the warped error scores at time $t$, and $\mathbbm{1}_{i=j}$ indicates if the warped error score $i$ is mapped to the pixel location $j$. The $\alpha$, $r_i$, $b_i$, and $s_i$ are the same as in the pixel color fusion process of Eq.~\ref{eq:m2m}. Note that since each  $e_i\in[0,1]$, the splatted error map $E_t(j)\in[0,1]$.

\begin{figure}[t]
    \centering
    \includegraphics[width=1\linewidth]{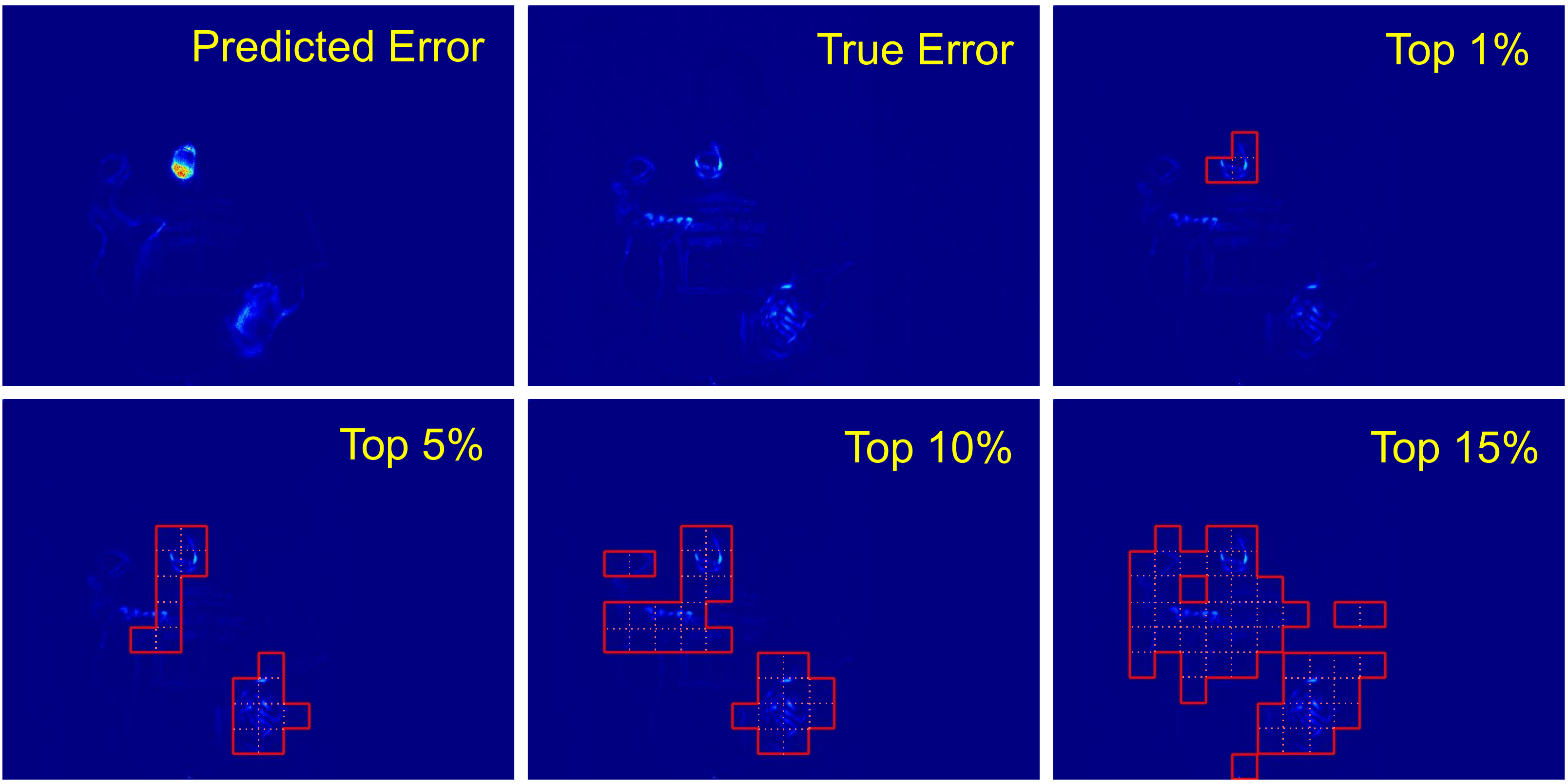}
    \vspace{-0.6cm}
    \caption{An example of the candidate patch selection step. Candidate patches are selected in a descend order of their predicted error scores. As shown, our error prediction is aligned with the true errors, and a larger  selection ratio  covers more inferior areas while requiring higher computational cost for further refinement.}
    \label{fig:patch_selection}
    \vspace{-0.3cm}
\end{figure}

So far, by applying many-to-many splatting, we can generate the interpolated video frame $I_{t}$ as well as the corresponding error prediction map $E_{t}$ where the error score of each target pixel is constituted by its associated motion vectors' error scores in an end-to-end differentiable way. To train the network to predict error scores for each motion vector, we minimize the L1 loss between the estimated error and actual interpolation residual for each pixel at the desired time step,
\begin{align}
\label{eq:errloss}
L_{Err} = |E_{t}-\frac{R_{t}}{R_{t}^{Max}}|
\end{align}
where $E_{t}$ is the error prediction, $R_{t}=\sum_{c\in\{R,G,B\}}|I_t^{c}-I^{c}_{GT}|$ is the absolute residual between the interpolation $I_t$ and the true $I_{GT}$, and $R_{t}^{Max}$ is the maximum value in $R_{t}$. The second term in Eq.~\ref{eq:errloss} normalizes the absolute residual to be between 0 and 1, facilitating the learning of $E_t$ which also varies between 0 and 1. A visualization of the error prediction process is illustrated in Fig.~\ref{fig:errorpred}.

\subsection{Candidate Patch Selection}
\label{subsec:cad_sel}
Given the error prediction map $E_t$, we downsample it to $\frac{1}{K}$ of its original resolution by max pooling and have a smaller error map $E_t'$, where each pixel represents a $K\times K$ patch of the original resolution. Then, pixels corresponding to the top-$p$ highest prediction errors in $E_t'$ are located and the corresponding $p$ patches in $E_t$ are selected for further refinement. 
By varying the value of $p$, different numbers of patches will be improved with fine-grained refinement, thus making it possible to trade computational efficiency for interpolation quality and vice versa. An example is shown in Fig.~\ref{fig:patch_selection}

\subsection{Patch Refinement Network}
\label{subsec:refine_net}
\begin{figure}[t]
    \centering
    \includegraphics[width=1\linewidth]{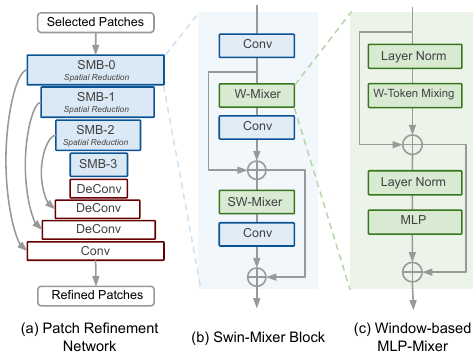}
    \vspace{-0.6cm}
    \caption{Architecture of our Patch Refinement Network (PRN). (a) PRN adopts an encoder-decoder structure with skip-connections. The encoder consists of (b) Swin-Mixer blocks that are based on W-Mixer and SW-Mixer, which are (c) Window-based MLP-Mixers with regular and shifted windowing configurations respectively.
    }
    \label{fig:patch_refinenet}
    \vspace{-0.3cm}
\end{figure}

We design our Patch Refinement Network (PRN) based on recent work on Swin Transformers~\cite{liu2021swin,lu2022vfitrans,liang2021swinir} as well as MLP-Mixers~\cite{tolstikhin2021mlp}. Specifically, our PRN adopts an encoder-decoder architecture as summarized in Fig.~\ref{fig:patch_refinenet} (a).
In the following, we subsequently describe the individual components in turn.

\subsubsection{Contextual Feature Pyramid.}
We follow \cite{niklaus2020softmax} and extract a feature pyramid from each input frame before warping these features to form a multi-resolution representation of the interpolation result. These warped feature pyramids then provide rich contextual features to the refinement network. However, instead of one-to-many splatting we perform our proposed many-to-many splatting to warp the pyramids. Furthermore, instead of executing the refinement network on the entire resolution, we only refine patches where the error prediction map is sufficiently high.
To achieve this, we first extract feature pyramids that will be used as inputs for PRNet. With the input pair $\{I_0, I_1\}$, we individually apply a CNN-based encoder to extract two four-level contextual feature pyramids $\{Q_0^i, Q_1^i\}_{i=0}^3$, with the image pair being the $0$-th level feature. Then, with the N full-resolution bidirectional motion fields $\{F_{0\rightarrow 1}^n,F_{1\rightarrow 0}^n\}_{n=1}^N$ estimated by the motion refinement network, we apply the many-to-many splatting for each direction separately and get two warped feature pyramids $\{Q_{t\_0}^i,Q_{t\_1}^i\}_{i=0}^3$ for a target time step $t$. Finally, for a patch $q$ selected from the error prediction map, we crop the corresponding patches from each level of $\{Q_{t\_0}^i,Q_{t\_1}^i\}_{i=0}^3$ and generate the patch-level contextual feature pyramids $\{q_{t\_0}^i,q_{t\_1}^i\}_{i=0}^3$.

\subsubsection{Swin-Mixer Block.} As shown in Fig.~\ref{fig:patch_refinenet} (b), each SMB consists of convolutional layers as well as two residual blocks. These two successive residual blocks are built on the Window-based MLP-Mixer (W-Mixer), with the second W-Mixer using a shifted window configuration~\cite{liu2021swin}.
When refining a patch, each SMB block not only gets the output from the preceding SMB block as input but also the warped contextual features from the corresponding level in the feature pyramid.


\subsubsection{Window-based MLP-Mixer.}
The window-based MLP-Mixer (W-Mixer) as illustrated in Fig.~\ref{fig:patch_refinenet} (c) is similar to a Swin transformer~\cite{liu2021swin} with the window-based Self-Attention operations replaced by window-based Token Mixing~\cite{tolstikhin2021mlp}. 
Given a feature map of size $K\times K\times C$ as the input, we evenly partition it into $\frac{K^2}{Q}$ non-overlapping sub-windows with size $\sqrt{Q}\times \sqrt{Q}$. 
For each window, its local feature map is flattened over the spatial domain, resulting in a two-dimensional feature table $X\in\mathcal{R}^{Q\times C}$. A token mixing MLP is applied to project $X$ along the
spatial dimension by,
\begin{align}
        U=X + W_2 \cdot \sigma(W_1 \cdot \text{LayerNorm}(X)) 
\end{align}
where $W_1\in\mathcal{R}^{M\times Q}$ and $W_2\in\mathcal{R}^{Q\times M}$ denote the weights of fully-connected layers, and $\sigma$ is the activation function GELU~\cite{hendrycks2016gaussian}. Then, the output is fed into subsequent layers that  project features along the channel dimension,
\begin{align}
        Y=U + \sigma(\text{LayerNorm}(U) \cdot W_1) \cdot W_2
\end{align}
where $W_1\in\mathcal{R}^{C\times D}$ and $W_2\in\mathcal{R}^{D\times C}$ denote the weights of fully-connected layers. Since the fixed window partition limits the information exchange across local windows, we apply regular and shifted window partitioning~\cite{liu2021swin} (with $\frac{\sqrt{Q}}{2}$ pixels) alternately in consecutive layers. As a result, the W-Mixer can efficiently aggregate features from each window and enables the Swin-Mixer Blocks to extract context from inputs in a hierarchical way.

\begin{table*}[t]
\caption{Quantitative results on the Vimeo90K, UCF101, ATD12K, and Xiph datasets. We compute models' GFLOPs and speed based on 640$\times$480 inputs. The ``share'' denotes the part of compute independent from the desired frame rate, which is in contrast to ``unshare''. In M2M++, we refine 100$\%$ patches for Vimeo90K, UCF101, as well as ATD20K, and the top 15$\%$ patches for Xiph.}
\vspace{-0.2cm}
\small
    \begin{tabular}{cccp{0.6cm}p{1.2cm}p{0.62cm}p{0.62cm}p{0.62cm}p{0.62cm}p{0.62cm}p{0.62cm}p{0.62cm}p{0.62cm}p{0.62cm}p{0.62cm}} 
        \toprule[1pt] 
                  & \multicolumn{2}{c}{GFLOPs} & \multirow{2}{0.4in}{Speed~~~\\~~ms/f~~~}  & \multirow{2}{0.4in}{Arbitrary~\\~~~Interp.}  & \multicolumn{2}{c}{Vimeo90K} & \multicolumn{2}{c}{UCF101} &\multicolumn{2}{c}{ATD12K} &\multicolumn{2}{c}{Xiph-2k} &\multicolumn{2}{c}{Xiph-``4k''} \\
                  \cmidrule(lr){2-3}\cmidrule(lr){6-7}\cmidrule(lr){8-9}\cmidrule(lr){10-11}\cmidrule(lr){12-13}\cmidrule(lr){14-15}
                  & \small{share}     &\small{unshare}      &   &     & \small{\footnotesize{PSNR}}        & \small{\footnotesize{SSIM}}         & \small{\footnotesize{PSNR}}        & \small{\footnotesize{SSIM}}       &\small{\footnotesize{PSNR}}        & \small{\footnotesize{SSIM}}        &\small{\footnotesize{PSNR}}        & \small{\footnotesize{SSIM}}        &\small{\footnotesize{PSNR}}        & \small{\footnotesize{SSIM}}       \\
\midrule
\footnotesize{SepConv\cite{niklaus2017sepconv}}     &\scriptsize{N/A}  &93              &~101           &                &33.79      &.970      &34.78      &.967      &27.40      &.950      &34.77       &.929      &32.06      &.880      \\
\footnotesize{DAIN\cite{bao2019depth}}              &712               &1308            &~977           &~~~$\checkmark$ &34.71      &.976      &35.00      &.968      &27.38      &.955      &35.97       &.940      &33.51      &.898      \\
\footnotesize{CAIN\cite{choi2020channel}}           &\scriptsize{N/A}  &29              &~~47           &                &34.65      &.973      &34.98      &.969      &25.28      &.952      &35.21       &.937      &32.56      &.901      \\
\footnotesize{AdaCoF~\cite{lee2020adacof}}          &\scriptsize{N/A}  &117             &~~36           &                &34.47      &.973      &34.90      &.968      &27.75      &.950      &34.82       &.927      &32.19      &.882      \\
\footnotesize{SoftSplat~\cite{niklaus2020softmax}}  &95                &218             &~122           &~~~$\checkmark$ &36.10      &.980      &\bf{35.39} &.970      &28.22      &.957      &36.62       &.944      &33.60      &.901      \\
\footnotesize{BMBC~\cite{park2020bmbc}}             &441               &376             &1213           &~~~$\checkmark$ &35.01      &.976      &35.15      &.969      &27.68      &.945      &~~~\scriptsize{--}      &~~~\scriptsize{--}     &~~~\scriptsize{--}      &~~~\scriptsize{--}     \\
\footnotesize{RIFE~\cite{huang2020rife}}            &\scriptsize{N/A}  &\underline{20}  &~~\textbf{17}  &                &35.51      &.978      &35.25      &.969      &28.59      &.953      &36.15       &.962      &33.27      &.942      \\
\footnotesize{ABME~\cite{park2021asymmetric}}       &\scriptsize{N/A}  &549             &~497           &                &36.18      &\bf{.981} &\udl{35.38}&.970      &28.71      &.959      &35.18       &.964      &32.36      &.940      \\
\footnotesize{FILM~\cite{reda2022film}}        &\scriptsize{N/A}  &341             &~~269$^*$           &          &36.06      &.970      &35.32      &.952       &--        &--        &36.66       &.951      &33.78      &.906      \\
\midrule
\footnotesize{RAFT-M2M}                &167                &\bm{$<1$}       &~~83           &~~~$\checkmark$             &35.51      &.978     &35.30       &.969      &\udl{29.41}&\udl{.960} &36.55      &.966      &33.91      &.944\\
\footnotesize{PWC-M2M}                 &87                &\bm{$<1$}       &~~32           &~~~$\checkmark$              &35.40      &.978      &35.17      &.970      &29.03      &.959       &36.45      &\udl{.967}&33.93      &.945\\
\footnotesize{DIS-M2M}                 &\textbf{61}       &\bm{$<1$}       &~~\underline{28}&~~~$\checkmark$             &35.06      &.976      &35.13      &.968      &28.95      &.956       &36.14      &.965      &33.25      &.942      \\
\midrule
\footnotesize{RAFT-M2M++}              &176               &357       &~351                 &~~~$\checkmark$              &\bf{36.20} &.980      &35.37      &.970      &\tbf{29.53}&\tbf{.961} &\udl{36.79}&.967      &\tbf{34.28} &\udl{.946}\\
\footnotesize{PWC-M2M++}             &96                &357             &~286           &~~~$\checkmark$              &\udl{36.19}&.980      &35.34      &.\tbf{970} &29.09     &.958       &\tbf{36.81}&\tbf{.968}&\udl{34.17} &\tbf{.947}\\
\footnotesize{{DIS-M2M++}}             &\underline{70}    &357             &~278           &~~~$\checkmark$              &35.78      &.980      &35.31      &.969       &29.00     &.958       &36.30      &.966      &33.47       &.942      \\
\bottomrule
    \end{tabular}
    \vspace{-0.2cm}
\label{tab:benchmark_s}
\end{table*}


\section{Experiments}
In the section, we compare our proposed approach to related state-of-the-art frame interpolation techniques and analyze it quantitatively as well as qualitatively.
\subsection{Datasets}
We supervise our approach only on the training split of Vimeo90K and test it on various datasets including: 1)~Vimeo90K~\cite{xue2019video}, the test split containing 3,782 triplets at a resolution of 448$\times$256 pixels. 2)~UCF101~\cite{soomro2012ucf101}, a dataset containing human action videos of size 256$\times$256 pixels. A set of 379 triplets were selected by Liu~\textit{et al.}~\cite{liu2017video} as a test set for frame interpolation. 3)~Xiph~\cite{xphi1994}, as proposed by Niklaus~\textit{et al.}~\cite{niklaus2020softmax} where ``Xiph-2K'' is generated by downsampling 4K footage, and ``Xiph-4k'' is based on center-cropped 2K patches. 4)~ATD12K~\cite{li2020video}, containing 2,000 triplets from various animation videos at a resolution of 960$\times$540 pixels. 5)~X-TEST~\cite{sim2021xvfi}, the test set from X4K1000FPS~\cite{sim2021xvfi}, containing 15 scenes extracted from 4K videos at 1000fps. We denote the original resolution as X-TEST(4K), and adopt X-TEST(2K) by downsampling X-TEST(4K) by a factor of two.

\subsection{Training}
We train the proposed M2M++ pipeline in two stages: first the many-to-many splatting framework, and then the Patch Refinement Network that further refines the initial interpolation results. The training loss for the many-to-many splatting framework consists of the sum of the Charbonnier loss~\cite{charbonnier1994two} as well as the census loss~\cite{meister2018unflow} for interpolated frame reconstruction, and the error estimation loss as defined in Eq.~\ref{eq:errloss} for predicting interpolation confidence. We initialize PWCNet with parameters pre-trained on the FlyingChairs dataset~\cite{dosovitskiy2015flownet}, and doesn't apply any auxiliary losses during training. To train the many-to-many splatting framework, we utilize the 51,312 triplets from the training split of Vimeo90K~\cite{xue2019video}. 
We apply random data augmentations including spatial and temporal flipping, color jittering, and random cropping with 256$\times$256 patches.
We adopt Adam~\cite{loshchilov2018fixing} for optimization, with a weight decay of 1e-4. 
We train the model for 800k iterations with a batch size of 8, during which the learning rate is decayed from 1e-4 to 0 via cosine annealing. 
After training the many-to-many splatting framework, we optimize the Path Refinement Network with similar hyperparameters for 400k iterations. Specifically, we adopt the combination of the Charbonnier loss~\cite{charbonnier1994two} and the census loss~\cite{meister2018unflow} for interpolation refinement, and train the network for 400k iterations with the sample data augmentation and learning rate scheduling strategy. 
Experiments are implemented with PyTorch, and evaluated on a Nvidia Titan X.

\subsection{Comparison with State-of-the-art}
We report multiple variants of our proposed approach based on different methods for estimating the off-the-shelf motion vectors. 
``PWC-M2M'' and ``PWC-M2M++'' are based on PWC-Net~\cite{sun2018pwc}, where we jointly optimize PWC-Net during training and generate initial flows at 1/4 of the original resolution. 
And analogously, ``DIS-M2M'' as well as ``DIS-M2M++'' are based on  DISFlow~\cite{kroeger2016fast}, ``RAFT-M2M'' as well as ``RAFT-M2M++'' are based on  RAFT-small~\cite{teed2020raft}.
In our experiments, we generate $N$=4 sub-motion vectors to many-splat each pixel.
For comparisons we report the performance of recent VFI approaches including: SepConv~\cite{niklaus2017sepconv}, DAIN~\cite{bao2019depth} CAIN~\cite{choi2020channel}, AdaCoF~\cite{lee2020adacof}, SoftSplat~\cite{niklaus2020softmax}, BMBC~\cite{park2020bmbc}, RIFE~\cite{huang2020rife},
ABME~\cite{park2021asymmetric}, and FILM~\cite{reda2022film}.

\begin{figure}[t]
\centering
    \includegraphics[width=1.0\linewidth]{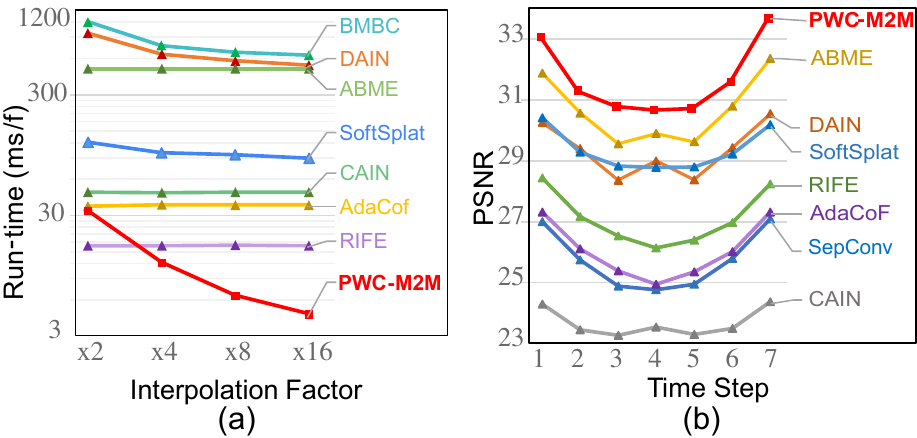}
    \vspace{-0.7cm}
    \caption{Evaluating multi-frame interpolation. (a) Runtime in logarithmic scale for interpolating 640$\times$480 video frames with different interpolation factors. (b) Per-frame accuracy for $\times$8 interpolation on X-TEXT(2K). Best viewed in color.}
    \vspace{-0.5cm}
    \label{fig:fr_vs_time}
\end{figure}

\begin{figure}
    \centering
    \includegraphics[width=0.92\linewidth]{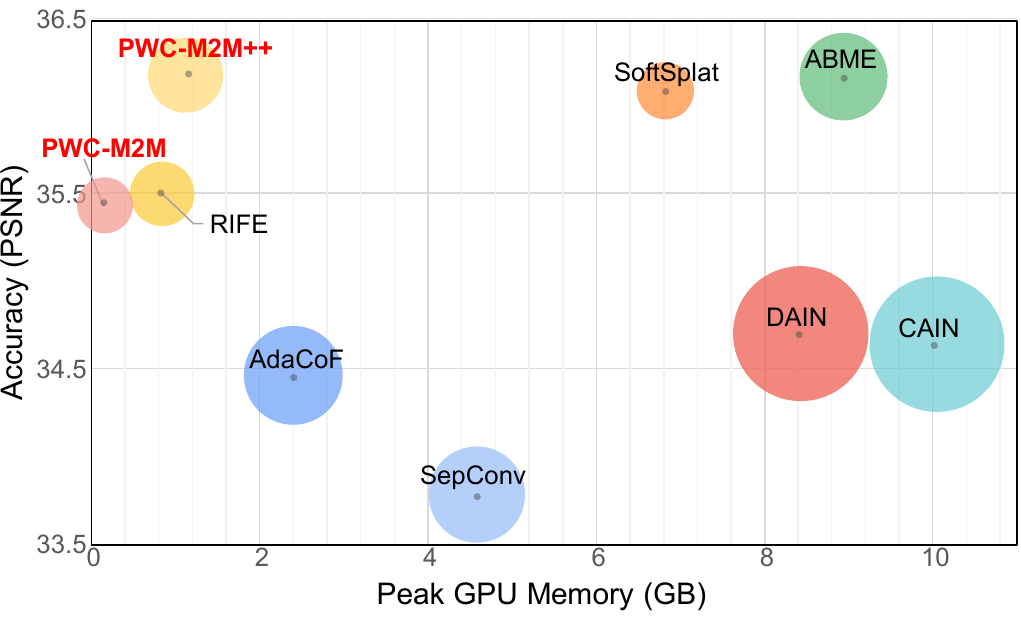}
    \vspace{-0.5cm}
    \caption{\small{ Performance comparison on Vimeo90K.The size of each circle indicates the number of model parameters.}}
    \vspace{-0.5cm}
    \label{fig:param_mem}
\end{figure}

We first analyze the computational efficiency of these models in Tab.~\ref{tab:benchmark_s}. 
We denote the required compute that is independent from the desired frame rate as ``share'', and  ``unshare''  otherwise.  
Hence the total computational complexity for interpolating $n$ frames can be calculated through ``$share+n\cdot unshare$''.
Motion-free methods (including SepConv, CAIN, and AdaCof) and pure bilateral-motion-based methods (like RIFE and ABME) have no shared compute (denoted as ``N/A'') and their computational complexity increases linearly in the number of desired frames. 
Approaches like SoftSplat, and BMBC can interpolate arbitrary frames, yet still suffer from both high unhsarable and sharable compute. For example in the $\times$8 interpolation setting,  they take 1.6 TFLOPs and 3.1 TFLOPs, respectively.
In contrast, our M2M takes only 0.1 TFLOPs in total.
Fig.~\ref{fig:fr_vs_time} (a) compares the average runtime for different methods subject to varying interpolation factors.  
Our M2M method is faster than all other methods in multi-frame settings. 
For $\times$16 interpolation our method takes about 5 ms to interpolate a frame, which is around 5$\times$, 20$\times$, and 100$\times$ faster than RIFE, SoftSplat, and ABME, respectively. 
Our M2M++ method introduces a post-interpolation refinement that incurs unsharable compute cost, yet as we can see from the table, the extra computational cost is still less than many of the previous methods like ABME etc. And in contrast to previous approaches with fixed computational costs, we can further improve the efficiency of M2M++ by refining fewer areas after the initial M2M interpolation. In Fig.~\ref{fig:param_mem}, we compare memory consumption and model size. As shown in the figure, our PWC-M2M requires the smallest model size and the lowest peak GPU memory, and our PWC-M2M++ achieves state-of-the-art accuracy with much less memory consumption than existing methods like ABME and SoftSplat.

\begin{table}[t]
\small
\caption{Quantitative results for $\times$8 interpolation on the X-TEST dataset. $^\dag$ indicates model trained with X-TRAIN. $^*$ indicates the numbers are copied from~\cite{park2021asymmetric}. All the run-times are measured on X-TEST(2K). For M2M++, we select the top 15$\%$ patches for refinement.}
\vspace{-0.3cm}
\begin{tabular}{cp{0.73cm}p{0.73cm}p{0.73cm}p{0.73cm}c}
\toprule[1pt] 
                                            & \multicolumn{2}{c}{X-TEST(4K)}        & \multicolumn{2}{c}{X-TEST(2K)} &Runtime\\
                                                \cmidrule(lr){2-3}\cmidrule(lr){4-6}
                                            & PSNR              & SSIM              & PSNR  & SSIM &(ms/f)\\
\midrule
SepConv~\cite{niklaus2017sepconv}           &23.94              &.794               &25.70          &.800      &~693         \\
DAIN~\cite{bao2019depth}                    &26.78$^{*}$        &.807$^{*}$         &29.33          &.910      &3132         \\
CAIN~\cite{choi2020channel}                 &22.51              &.775               &23.62          &.773      &~287         \\
AdaCoF~\cite{lee2020adacof}                 &23.90              &.727               &26.03          &.778      &~234         \\
SoftSplat~\cite{niklaus2020softmax}         &25.48              &.725               &29.73          &.824      &~318        \\
RIFE~\cite{huang2020rife}                   &24.67              &.797               &27.49          &.806      &~104         \\
ABME~\cite{park2021asymmetric}              &30.16$^{*}$        &.879$^{*}$         &30.65          &.912      &~2904         \\
XVFI$^{\dag}$~\cite{sim2021xvfi}            &30.12              &.870               &30.85          &.913      &~203         \\
\midrule
RAFT-M2M                                    &\udl{31.46}        &\udl{.922}         &32.12          &\udl{.931}&~~116\\
PWC-M2M                                     &30.81              &.912               &32.07          &.923      &~~\udl{44}\\
DIS-M2M                                     &30.18              &.909               &30.98          &.912      &~~\tbf{39}\\
\midrule
RAFT-M2M++                                  &\tbf{31.49}        &\tbf{.923}         &\tbf{32.41}    &\tbf{.932}&~402\\
PWC-M2M++                                   &30.94              &.914               &\udl{32.24}    &.924      &~314\\
DIS-M2M++                                   &30.24              &.911               &31.66          &0.922      &~307\\
\bottomrule
\end{tabular}
\vspace{-0.5cm}
\label{tab:benchmark_xvfi}
\end{table}

Taking efficiency aside, our method demonstrates its effectiveness on multiple datasets.
The metrics for $\times$2 interpolation are presented in Tab.~\ref{tab:benchmark_s}. 
On Vimeo90K and UCF101, our M2M method is on par with the recently proposed real-time method RIFE and performs slightly worse than SoftSplat and ABME.
On Xiph-2K, our M2M method achieves slightly lower PSNR than SoftSplat, yet achieves the highest SSIM among all the methods. 
Moreover, on the animation dataset ATD12K and the high-resolution dataset Xiph-``4K'', our M2M method, especially PWC-M2M, outperforms previous methods in terms of both PSNR and SSIM. 
This demonstrates our M2M method's effectiveness when processing high-resolution videos and the ability to generalize across domains such as animation videos.
With the selective refinement step, our M2M++ can further improve performance. On Vimeo90K and UCF101, M2M++ achieves comparable accuracy to state-of-the-art methods like ABME, while being computationally much more efficient. Compared to the recent FILM, our PWC-M2M++ runs at a similar speed yet achieves consistently better accuracy on different  datasets. And on datasets like  ATD12K and Xiph, M2M++ further boosts the performance and outperforms previous methods with significant margin.

We report  the results for $\times$8 interpolation on the X-TEST dataset, which contains diverse sequences with both high resolution and high frame rate, in Tab.~\ref{tab:benchmark_xvfi}. 
Our M2M method outperforms all previous methods on both the original 4K full resolution (4096$\times$2160) and the downsampled 2K resolution (2048$\times$1080) with substantial advantages in efficiency. And by further refining only 15$\%$ of the image regions, our M2M++ method successfully achieves new state-of-the-art performance.
For the models trained with Vimeo90K, ABME achieves the second-best PSNR in both 4K and 2K settings, but it takes 2,904ms to interpolate a 2K frame which is nearly 70$\times$ slower than M2M.
To evaluate the quality of multi-frame interpolation, we also compare step-wise accuracy for $\times8$ interpolation in Fig.~\ref{fig:fr_vs_time} (b).
We found that previous methods tend to deteriorate when interpolating frames that are temporally centered between the inputs. In contrast, M2M can more stably and accurately generate intermediate frames at arbitrary time steps.
\subsection{Analysis of M2M}
\begin{table}[]
\centering
\caption{Ablative experiments (in PSNR) on Vimeo90K with different initial flow methods. ``MRN'' denotes the motion refinement network, ``JFE'' refers to the joint flow encoding module in MRN, ``LFM'' is the low-rank feature modulation, and ``RS'' denotes the reliability score in the fusion step that synthesizes the output.}
\vspace{-0.4cm}
\begin{tabular}{p{0.77cm}p{0.77cm}p{0.77cm}p{0.77cm}|cc}
\toprule[1pt] 
                          {MRN}  &{JFE} &{LFM}       &{RS}   &{PWC-Net} &{DISFlow} \\ 
    \hline
               &               &                   &                               &33.97    &31.93\\ 
    \hline
  ~~~$\checkmark$ &               &                   &                               &34.94    &34.32\\ 
  ~~~$\checkmark$ &~~~$\checkmark$   &                   &                               &35.09    &34.59\\ 
  ~~~$\checkmark$ &               &~~~$\checkmark$       &                               &35.07    &34.51\\ 
  \hline
  ~~~$\checkmark$ &~~~$\checkmark$   &~~~$\checkmark$       &                               &35.15    &34.78\\ 
  ~~~$\checkmark$ &~~~$\checkmark$   &~~~$\checkmark$       &~~~$\checkmark$                   &35.24    &34.93\\ 
\bottomrule
\end{tabular}
\label{tab:ablation}
\vspace{-0.3cm}
\end{table}
\begin{table}[t]
\centering
\caption{Analyzing the impact of the number of sub-motion vectors for each pixel in our many-to-many splatting on Vimeo90K, with two different initial flow estimators.}
\vspace{-0.4cm}
\renewcommand\arraystretch{1.2} 
\begin{tabular}{cc|cccc}
\toprule[1pt] 
                                &                       &$N$=1      &$N$=2      &$N$=4  &$N$=8\\ 
    \hline
    \multirow{3}{*}{\small{PWC-Net} }   &\small{PSNR}           &35.24     &35.35       &35.40  &35.39\\ 
                                &\small{Runtime}        &16        &16          &17     &20       \\ 
                                &\small{GFLOPs}        &36.1        &36.2          &36.4     &36.7       \\ 
                                 \hline
    \multirow{3}{*}{\small{DISFlow} }   &\small{PSNR}           &34.93     &34.98       &35.06  &35.07\\ 
                                &\small{Runtime}        &16        &16          &17     &20       \\ 
                                &\small{GFLOPs}        &26.0        &26.1          &26.3     &26.6       \\ 
\bottomrule[1pt]
\end{tabular}
\vspace{-0.3cm}
\label{tab:branch}
\end{table}

\subsubsection{Ablation of Modules.} We first analyze the effectiveness of the different components of our method in Tab.~\ref{tab:ablation}. 
We start with a single motion vector for each pixel.
The first row demonstrates that directly using the off-the-shelf flow for warping leads to sub-optimal accuracy.
As shown in the second row,  applying the refinement network without joint flow encoding (JFE) and low-rank feature modulation (LFM) can already significantly improve performance by 0.97 dB and 2.38 dB for PWC-Net and DISFlow, respectively.
Further applying either JFE or LFM leads to improvements of more than  0.15 dB for both off-the-shelf flow methods.
And using both JFE and LFM helps to boost the performance to 35.15 dB and 34.78 dB, respectively.
In the last two rows, we also show the impact of the reliability scores which are generated by the refinement network and utilized for the pixel fusion. 
Without this score, the performance degrades, thus highlighting the importance of this metric in comparison to only using photoconsistency.

\begin{figure*}[t]
    \centering
    \includegraphics[width=1\linewidth]{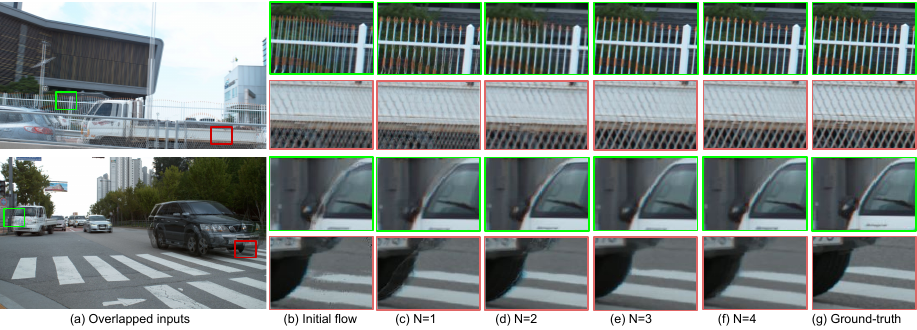}
    \vspace{-0.7cm}
    \caption{Analysis of  many-to-many splatting. Given the input frames (a), M2O splatting with the initial flow (b) or single refined sub-motion vector (c) results in undesired visual artifacts for regions with complex motion. In comparison, M2M splatting with more sub-motion vectors (d)-(f) can interpolate with higher quality.}
    \vspace{-0.4cm}
    \label{fig:xtest_vis}
\end{figure*}
\subsubsection{Effect of Number of Flows per Pixel.}
Tab.~\ref{tab:branch} compares the effect of using different numbers of sub-motion vectors for the M2M splatting.
When $N$=1, it reduces the warping to many-to-one (M2O) splatting, and achieves the lowest accuracy.
When increasing $N$ to 4,  M2M improves the accuracy by more than 0.1 dB, with a very slight increment in run-time ($<$1ms) and complexity ($<$0.3 GFLOPs). 
Also, and as shown in the last row, we noticed that further increasing the number of sub-motion vectors leads to marginal improvements. As visualized in
Fig.~\ref{fig:xtest_vis}, compared to M2O splatting, the proposed M2M splitting helps to improve the interpolation quality of challenging areas like boundaries and complex motion.

\begin{table}[t]
\centering
\small
\caption{Impact of the resolution at which the initial optical flow estimator is applied on. ``R'' is the down-sampling factor.}
\vspace{-0.4cm}
\begin{tabular}{p{0.2cm}cccccc}
    \toprule[1pt] 
    &\textit{R=}     &\footnotesize{Xiph-2K}  &\footnotesize{Xiph-``4k''}  &\footnotesize{X-TEST(2K)}  &\footnotesize{X-TEST(4K)} \\
    \midrule
    \multirow{4}{*}{\rotatebox{90}{PWC-Net} }&1   &36.15              &32.94           &28.35             &24.85\\           
                                             &2   &\textbf{36.45}     &33.76           &31.00             &27.08\\           
                                             &4   &36.36              &\textbf{33.93}  &\textbf{32.07}    &29.65\\           
                                             &8   &35.74              &33.75           &31.65      &\textbf{30.81}\\
    \midrule
    \multirow{4}{*}{\rotatebox{90}{DISFlow} }&1   &\textbf{36.14}     &\textbf{33.25}  &31.03             &\textbf{30.18}\\           
                                             &2   &36.05              &33.18           &\textbf{31.18}    &30.06\\           
                                             &4   &35.73              &32.94           &30.54             &29.68\\           
                                             &8   &35.13              &32.29           &29.49             &28.66\\ 
    \bottomrule[1pt]
\end{tabular}
\vspace{-0.3cm}
\label{tab:resolution}
\end{table}

\subsubsection{Effect of Resolution for Initial Flow Estimation.}
Our method relies on an off-the-shelf optical flow estimator to generate the initial flow. 
However, most optical flow estimation models are trained using a relatively low resolutions.
Directly applying them to estimate the flow at 2K or 4K inputs  may hence result in sub-optimal results.
We thus study the impact of the initial flow's resolution for interpolating high-resolution frames in Tab.~\ref{tab:resolution}. Since PWC-Net is learning-based and pre-trained on small resolutions, it is less effective at processing high-resolution frames as demonstrated by the reduced interpolation quality on 4K data.
By downsampling the input by a factor of 4 or 8, the accuracy improves significantly.
In contrast, DISFlow is not supervised and hence less susceptible to similar domain gaps. 

\begin{table}[t]
\centering
\caption{Comparisons between PWCNet-based M2M and M2M++ with similar computational complexities. ``M2M$^{\times n}$'' denotes M2M with intermediate feature channels increased by $n$ times. ``M2M$++^{n\%}$'' indicates M2M++ with refinement ratio ${n\%}$.}
\vspace{-0.4cm}
\begin{tabular}{p{0.93cm}<{\centering}|p{0.73cm}<{\centering}|p{0.95cm}<{\centering}p{1.4cm}<{\centering}|p{0.95cm}<{\centering}p{1.4cm}<{\centering}}
    \toprule
    &M2M &M2M$^{\times2}$&M2M++$^{50\%}$&M2M$^{\times2.5}$&M2M++$^{100\%}$\\
    \hline
    PSNR    &35.40  &35.72  &36.05  &35.81  &36.19\\
    GFLOPs  &37     &141    &148    &253    &261\\
    \bottomrule
\end{tabular}
\label{tab:complexity_psnr}
\vspace{-0.3cm}
\end{table}
\begin{figure}[]
    \centering
    \includegraphics[width=0.85\linewidth]{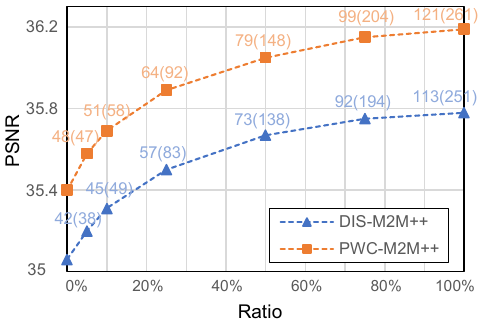}
    \vspace{-0.4cm}
    \caption{Performance of M2M++ with different refinement ratio on Vimeo90K. The numbers above each node represent the corresponding interpolation speed (ms/f) and GFLOPs (in the brackets).} 
    \vspace{-0.4cm}
    \label{fig:ratio_perf}
\end{figure}

\begin{table}[t]
\centering
\caption{Analysis for ratio of remaining empty pixels in many-to-many splatting.}
\vspace{-0.4cm}
\begin{tabular}{ccccc}
    \toprule
                        &Vimeo90K   &ATD20K     &Xiph-2K    &XTEST-4K\\
    \midrule
        Initial Flow    &0.0047$\%$     &0.0067$\%$     &0.0008$\%$    &0.0135$\%$\\
        M2M (N=4)       &0.0006$\%$     &0.0003$\%$     &0.0001$\%$    &0.0004$\%$\\
    \bottomrule
\end{tabular}
\label{tab:hole}
\vspace{-0.3cm}
\end{table}

\begin{figure}[t]
    \centering
    \includegraphics[width=\linewidth]{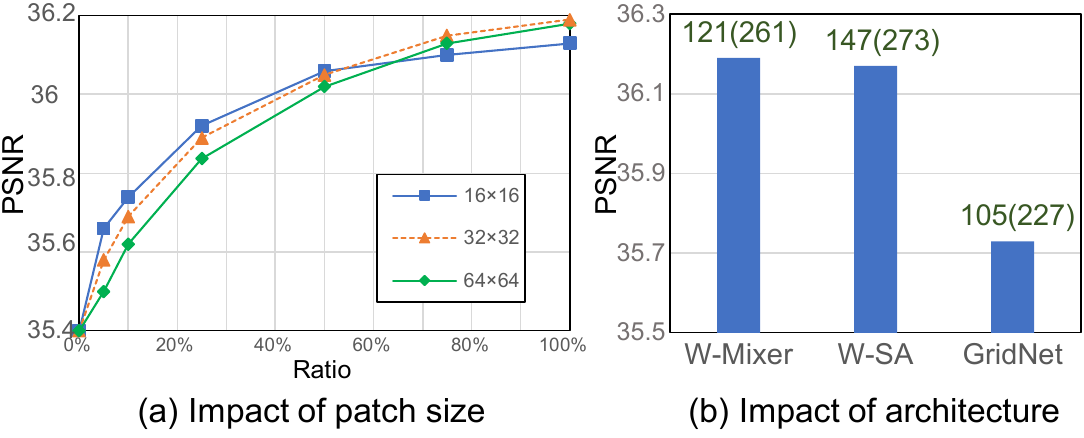}
    \vspace{-0.4cm}
    \caption{Analysis for the architecture designs of Patch Refinement Network. The numbers on each bar represent the corresponding interpolation speed (ms/f) and GFLOPs (in the brackets).}
    \vspace{-0.4cm}
    \label{fig:arh_ana}
\end{figure}
\begin{figure*}[t]
    \centering
    \includegraphics[width=1\linewidth]{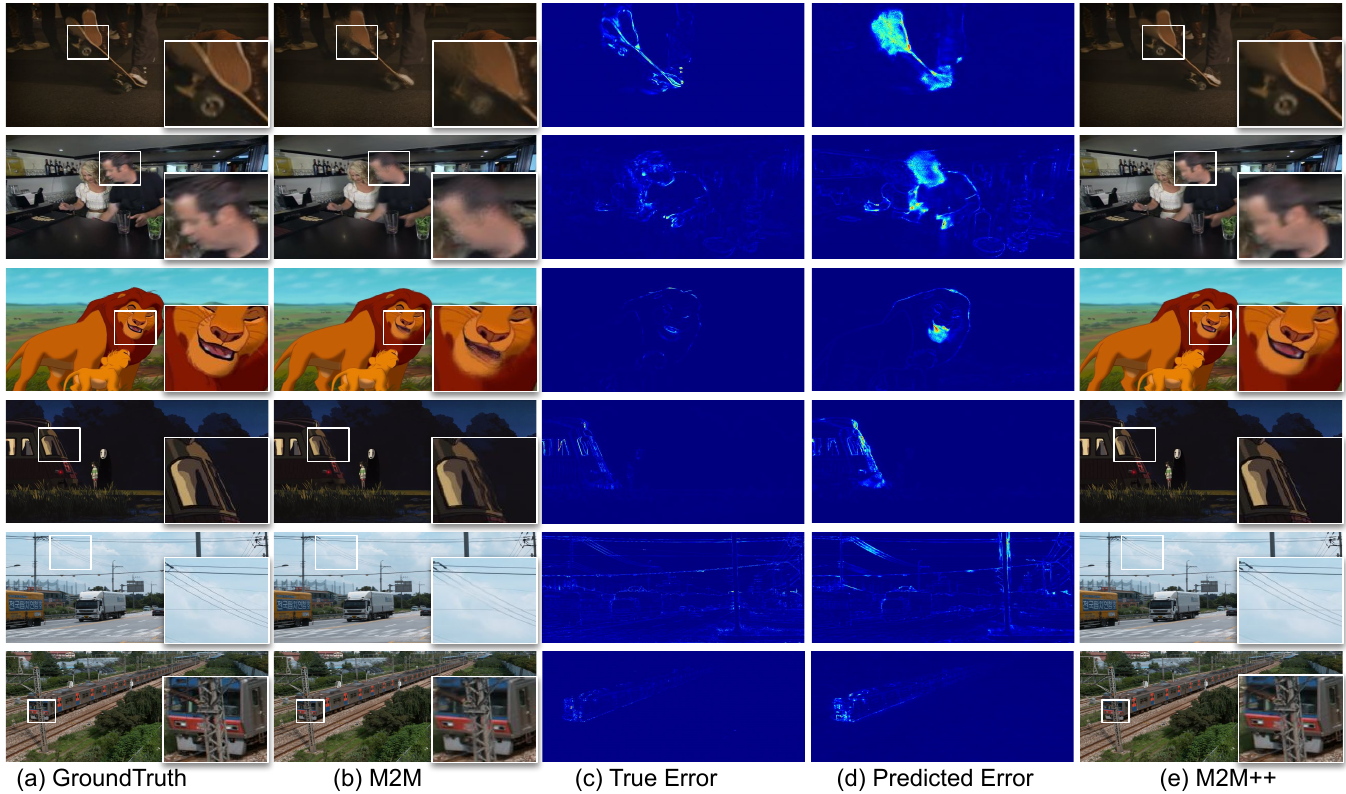}
    \vspace{-0.8cm}
    \caption{Representative qualitative results. The predicted error maps are aligned with the true error maps of initial interpolation, and M2M++ is able to rectify the undesired effects like blurry, distortion, etc caused by the many-to-many splatting of pixel colors.}
    \label{fig:qual_result}
    \vspace{-0.5cm}
\end{figure*}

\subsubsection{Impact of Model Size.}
In Tab.~\ref{tab:complexity_psnr}, we enlarge M2M's model capacity to analyze the impact on performance. We increase the intermediate feature channels in M2M by 2 times and 2.5 times to match the computational complexity of M2M++ under 50$\%$ and 100$\%$ refinement respectively. As shown in the table, enlarging M2M's model size helps to further increase the accuracy. However, it is still less effective than M2M++ with similar computational complexity. We also replace PWCNet with RAFT for the initial optical flow estimation. As compared in Tab.~\ref{tab:benchmark_s} and Tab.~\ref{tab:benchmark_xvfi}, RAFT-M2M takes about 2 times of GFLOPs compared to PWC-M2M, yet improves performance on most of the datasets. And by applying the Selective Refinement method, RAFT-M2M++ further boosts the state-of-the-art performance, especially on high-resolution datasets like ATD12K and XTEST. This demonstrates the impact of model capacity and the effectiveness of M2M++.

\subsubsection{Hole Pixels.}
Though M2M achieves very high efficiency, especially for high frame rate interpolation, it renders intermediate frames based on forwarding warping which may be subject to holes in the output. 
We analyze the number of holes as a percentage of the output resolution on various datasets in Tab.~\ref{tab:hole}.
With 0.0006$\%$ of the area being empty on Vimeo90K, our M2M with four flow fields generates only 0.7 empty pixels per image, which is 8$\times$ better than the single initial flow-based warping. We also observe that M2M improves the ratio of hole pixels by more than 20$\times$ on ATD20K and XTEST-4K. This demonstrates that our many-to-many splatting can help to reduce the number of empty pixels.


\subsection{Analysis of M2M++}
\subsubsection{Speed-accuracy Trade-off.} 
After many-splatting a given input frame pair, we are able to get an initial interpolation result and the corresponding error prediction map. Based on the error prediction map, we rank patches in the interpolation result based on their error estimates. By varying the selection ratio, it is possible to refine fewer patches to improve the computational efficiency or refine more patches to improve the interpolation quality. We demonstrate this ability to trade computational efficiency for interpolation quality and vice versa in Fig.~\ref{fig:ratio_perf}. To summarize, a larger ratio always leads to better interpolation quality at the cost of computational efficiency but the improvement in the large ratio area (e.g. from 75$\%$ to 100$\%$) is marginal compared to the small ratio area (e.g. from 0$\%$ to 25$\%$). To further validate the quality of the error prediction in M2M++, we replace the predicted error map with the true error and compare the performance in Tab.~\ref{tab:err_vs_gt}. In this experiment,  our predicted error performs very similarly to the true error which demonstrates that the predicted error maps can effectively locate areas with low-quality interpolation results and characterize the degree of error. Some examples are shown in Fig.~\ref{fig:qual_result}.

\begin{table}[t]
\centering
\small
\caption{Performance of PWC-M2M++ with predicted interpolation error and true interpolation error.}
\vspace{-0.4cm}
\begin{tabular}{cccccccc}
    \toprule
    Ratio=          &0$\%$   &5$\%$   &10$\%$   &25$\%$   &50$\%$   &75$\%$\\
    \hline
    True Error      &35.40 &35.60	&35.71	&35.90	&36.06	&36.15\\
    Pred. Error     &35.40 &35.58	&35.69	&35.89	&36.05	&36.15\\
    \bottomrule
\end{tabular}
\label{tab:err_vs_gt}
\vspace{-0.5cm}
\end{table}

\subsubsection{Impact of Patch Size.}
In the selective refinement step, patches are selected from the initial interpolation for further improvement. We analyze the impact of patch size with varied refinement ratios in Fig.~13 (a). We find that under a lower refinement ratio, the smaller patch sizes achieve higher accuracy, e.g. given Ratio=25$\%$, patch size 16$\times$16 achieves the highest PSNR while patch size 64$\times$64 performs the worst. This may be because a smaller patch is more flexible and can cover more non-contiguous candidate pixels given a low selection ratio. When increasing the refinement ratio to 75$\%$, we notice that the larger patch sizes start to perform better, due to the larger context contained in each patch.

\subsubsection{Impact of Swin-Mixer Block.}
To evaluate the effectiveness of the Window-Based MLP-Mixer (W-Mixer) in the Patch Refinement Network, we analyze the performance by replacing it with the Window-Based Self-Attention (W-SA). As shown in Fig.~\ref{fig:arh_ana} (b), W-SA achieves similar accuracy yet increases the inference time from  121 ms/f to 147 ms/f. We also compare our Patch Refinement Network with a GridNet~\cite{fourure2017residual,niklaus2018context}, and find that GridNet runs at a slightly faster speed but decreases accuracy by 0.5 PSNR compared to our PRN. The results demonstrate that our method designs attains a good balance between accuracy and efficiency.

\section{Conclusion}
In this work, we present a many-to-many splatting technique to efficiently interpolate intermediate video frames, and further extend this framework with a spatial selective refinement module to dynamically improve the interpolation quality at erroneous regions. 
Specifically, we propose a Motion Refinement Network to generate multiple sub-motion vectors for each pixel. 
These sub-motion fields are then applied to forward warp the pixels to any desired time step before fusing the splatted pixels to obtain the interpolation output.
By sharing the computation for the flow refinement and only requiring little computation to generate each frame, our method is especially well-suited for multi-frame interpolation.
Based on this fast initial interpolation, we further apply a spatial selective refinement to process certain regions selected with the guidance of a predicted error map, hence avoiding unnecessary computation.
This selective refinement allows trading computational efficiency for interpolation quality and vice versa.
Experiments on multiple benchmark datasets demonstrate that the proposed method combines effectiveness and efficiency.

\bibliographystyle{IEEEtran}
\bibliography{egbib}

\end{document}